\documentclass[english]{IEEEtran}
\usepackage[T1]{fontenc}
\usepackage[latin9]{inputenc}
\usepackage{amsmath}
\usepackage{graphicx}
\usepackage[numbers]{natbib}

\makeatletter
\IfFileExists{lmodern.sty}{\usepackage{lmodern}}{}
\usepackage{amsmath}
\DeclareMathOperator*{\argmax}{arg\,max}

\usepackage{xcolor}
\usepackage[pagebackref=true]{hyperref}
\hypersetup{
    colorlinks,
    linkcolor={red!50!black},
    citecolor={blue!50!black},
    urlcolor={blue!80!black}
}

\makeatother

\usepackage{babel}
\begin{document}

\title{Bridging Cognitive Programs and Machine Learning}

\author{Amir Rosenfeld, John K. Tsotsos  \\
Department of Electrical Engineering and Computer Science\\ York University, Toronto, ON, Canada\\
\texttt{amir@eecs.yorku.ca,tsotsos@cse.yorku.ca} \\
}
\maketitle

\section{abstract}
While great advances are made in pattern recognition and machine learning, the successes of such fields remain restricted to narrow applications and seem to break down when training data is scarce, a shift in domain occurs, or when intelligent reasoning is required for rapid adaptation to new environments. In this work, we list several of the shortcomings of modern machine-learning solutions, specifically in the contexts of computer vision and in reinforcement learning and suggest directions to explore in order to try to ameliorate these weaknesses.
\section{Introduction}

The Selective Tuning Attentive Reference (STAR) model of attention
is a theoretical computational model designed to reproduce and predict
the characteristics of the human visual system when observing an image
or video, possibly with some task at hand. It is based on psycho-physical
observations and constraints on the amount and nature of computations
that can be carried out in the human brain. The model contains multiple
sub-modules, such as the Visual Hierarchy (VH), visual working memory
(vWM), fixation controller (FC), and other. The model describes flow
of data between different components and how they affect each other.
As the model is given various tasks, an executive controller orchestrates
the action of the different modules. This is viewed as a general purpose
processor which is able to reason about the task at hand and formulate
what is called Cognitive Programs (CP). Cognitive Programs are made
up of a language describing the the set of steps required to control
the visual system, obtain the required information and track the sequence
of observations so that the desired goal is achieved. In recent years,
methods of pattern recognition have taken a large step forward in
terms of performance. Visual recognition of thousands of object classes
as well as detection and segmentation have been made much more reliable
than in the past. In the related field of artificial intelligence,
progress has been made by the marriage of reinforcement learning and
deep learning, allowing agents to successfully play a multitude of
game and solve complex environments without the need for manually
crafting feature spaces or adding prior knowledge specific to the
task. There is much progress still to be made in all of the above
mentioned models, namely 

(1) a computational model of the human visual system (2) purely computational
object recognition systems (e.g, computer vision) and (3) intelligent
agents. The purpose of this work is to bridge the gap between the
worlds of machine learning and modeling of the way human beings solve
visual tasks. Specifically, providing a general enough solution to
the problem of coming up with Cognitive Programs which will enable
solving visual tasks given some specification. 

We make two main predictions: 
\begin{enumerate}
\item Many components of the STAR model can benefit greatly from modern
machine learning tools and practices. 
\item Constraining the machine learning methods used to solve tasks, using
what is known on biological vision will benefit these models and,
if done right, improve their performance and perhaps allow us to gain
further insights.
\end{enumerate}
The next sections will attempt to briefly overview the STAR model
as well as the recent trends in machine learning. In the remainder
of this report, we shall show how the best of both worlds of STAR
and Machine Learning can be brought together to create a working model
of an agent which is able to perform various visual tasks.

\section{Selective Tuning \& Cognitive Programs}

The Selective Tuning (ST) \citep{tsotsos1993inhibitory,tsotsos1995modeling,culhane1992attentional,books/daglib/0026815}
is a theoretical model set out to explain and predict the behavior
of the human visual system when performing a task on some visual input.
Specifically, it focuses on the phenomena of visual attention, which
includes overt attention (moving the eyes to fixate on a new location),
covert attention (internally attending to a location inside the field
of view without moving the eyes) and neural modulation and feedback
that facilitates these processes. The model is derived from first
principles which involve analysis of the computational complexity
of general vision tasks, as well as biological constraints known from
experimental observation on human subjects. Following these constraints,
it aims to be biologically plausible while ensuring a runtime which
is practical (in terms of complexity) to solve various vision tasks.
In \citep{tsotsos2014cognitive} , ST has been extended to the STAR
(Selective Tuning Attentive Reference) model to include the capacity
for cognitive programs. 

We will now describe the main components of STAR. This description
is here to draw a high-level picture and is by no means complete.
For a reader interested in delving into further details, please refer
to \citep{books/daglib/0026815} for theoretical justifications and
a broad discussion and read \citep{tsotsos2014cognitive} for further
description of these components. The ST model described here is extended
with a concept of Cognitive Programs (CP) which allows a controller
to break down visual tasks into a sequence of actions designed to
solve them.
\begin{center}
\begin{figure*}
\begin{centering}
\includegraphics[angle=90,width=0.9\textwidth]{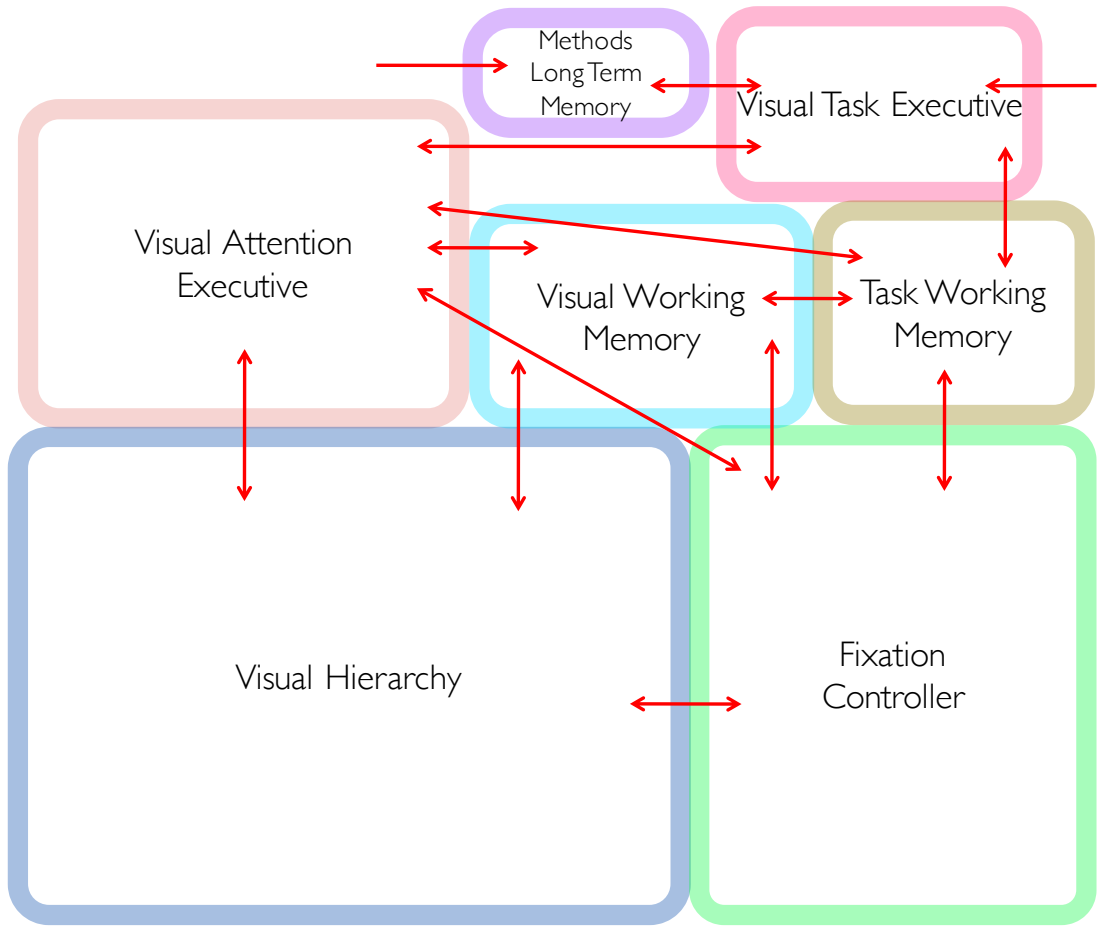}
\par\end{centering}
\caption{\label{fig:High-level-view-of}High-level view of the STAR architecture.
Reproduced from \citep{tsotsos2014cognitive}}
\end{figure*}
\par\end{center}

Fig. \ref{fig:High-level-view-of} describes the flow of information
in the STAR architecture at a high-level. Central to this architecture
is the Visual Hierarchy. The VH is meant to represent the ventral
and dorsal streams of processing in the brain and is implemented as
a neural network with feedforward and recurrent connections. The structure
of the VH is designed to allow recurrent localization of input stimuli,
as well as discrimination, categorization and identification. While
a single feed-forward pass may suffice for some of the tasks, for
other, such as visual search, multiple forward-backward passes (and
possibly changing the focus of attention) may be required. Tuning
of the VH is allowed so it will perform better on specific tasks.
The recurrent tracing of neuron activation along the hierarchy is
performed using a $\Theta$-WTA decision process. This induces an
Attentional Sample (AS) which represents the set of neurons whose
response matches the currently attended stimulus. 

The Fixation Control mechanism has two main components. The Peripheral
Priority Map (PPM) represents the saliency of the peripheral visual
field. The History Biased Priority Map (HBPM) combines the focus of
attention derived from the central visual field (cFOA) and the foci
of attention derived from the peripheral visual field (pFOA). Together,
these produce a map based on the previous fixations (and possibly
the current task), setting the priority for the next gaze. 

\subsection{Cognitive Programs}

To perform some task, the Visual Hierarchy and the Fixation Controller
need to be controlled by a process which receives a task and breaks
it down into a sequence of \emph{methods}, which are basic procedures
commonly used across the wide range of visual tasks. Each method may
be applied with some degree of tuning to match it to the specific
task at hand, whereas it become an executable \emph{script. }A set
of functional sub-modules is required for the execution of CP's.

The controller orchestrating the execution of tasks is called the
Visual Task Executive (vTE). Given a task (from some external source),
the vTE selects appropriate methods, tunes them into scripts and controls
the execution of these scripts by using several sub-modules. Each
script initiates an attentive cycle and sends the element of the task
required for attentive tuning to the Visual Attention Executive (vAE).
The vAE primes the Visual Hierarchy (VH) with top-down signals reflecting
the expectations of the stimulus or instructions and sets required
parameters. Meanwhile, the current attention is disengaged and any
feature surround suppression imposed for previous stimuli is lifted.
Once this is completed, a feed-forward signal enters the tuned VH.
After the feed-forward pass is completed, the $\Theta$-WTA process
selects the makes a decision as to what to attend and passes on this
choice from the next stage. The vTE, monitoring the execution of the
scripts, can decide based on this information whether the task is
completed or not. 

The selection of the basic methods to execute a task is done by using
the Long Term Memory for Method (mLTM). This is an associative memory
which allows for fast retrieval of methods. 

The Visual Working Memory (vWM) contains two representations: the
Fixation History Map stores the last several fixation locations, each
decaying over time. This allows for location based Inhibition of Return
(IOR). The second representation is the Blackboard (BB), which store
the current Attentional Sample (AS). 

Task Working Memory (tWM) includes the Active Script NotePad which
itself might have several compartments. One such compartment would
store the active scripts with pointers to indicate progress along
the sequence. Another might store information relevant to script progress
including the sequence of attentional samples and fixation changes
as they occur during the process of fulfilling a task. Another might
store relevant world knowledge that might be used in executing the
CP. The Active Script NotePad would provide the vTE with any information
required to monitor task progress or take any corrective actions if
task progress is unsatisfactory.

Finally, the Visual Attention Executive contains a Cycle Controller,
which is responsible for starting and terminating each stage of the
ST process. The vAE also initiates and monitors the recurrent localization
process in the VH \citep{rothenstein2014attentional}. A detailed
view of the entire architecure can be seen in Fig \ref{fig:Detailed-view-of}.
\begin{center}
\begin{figure*}
\begin{centering}
\includegraphics[angle=90,width=0.9\textwidth]{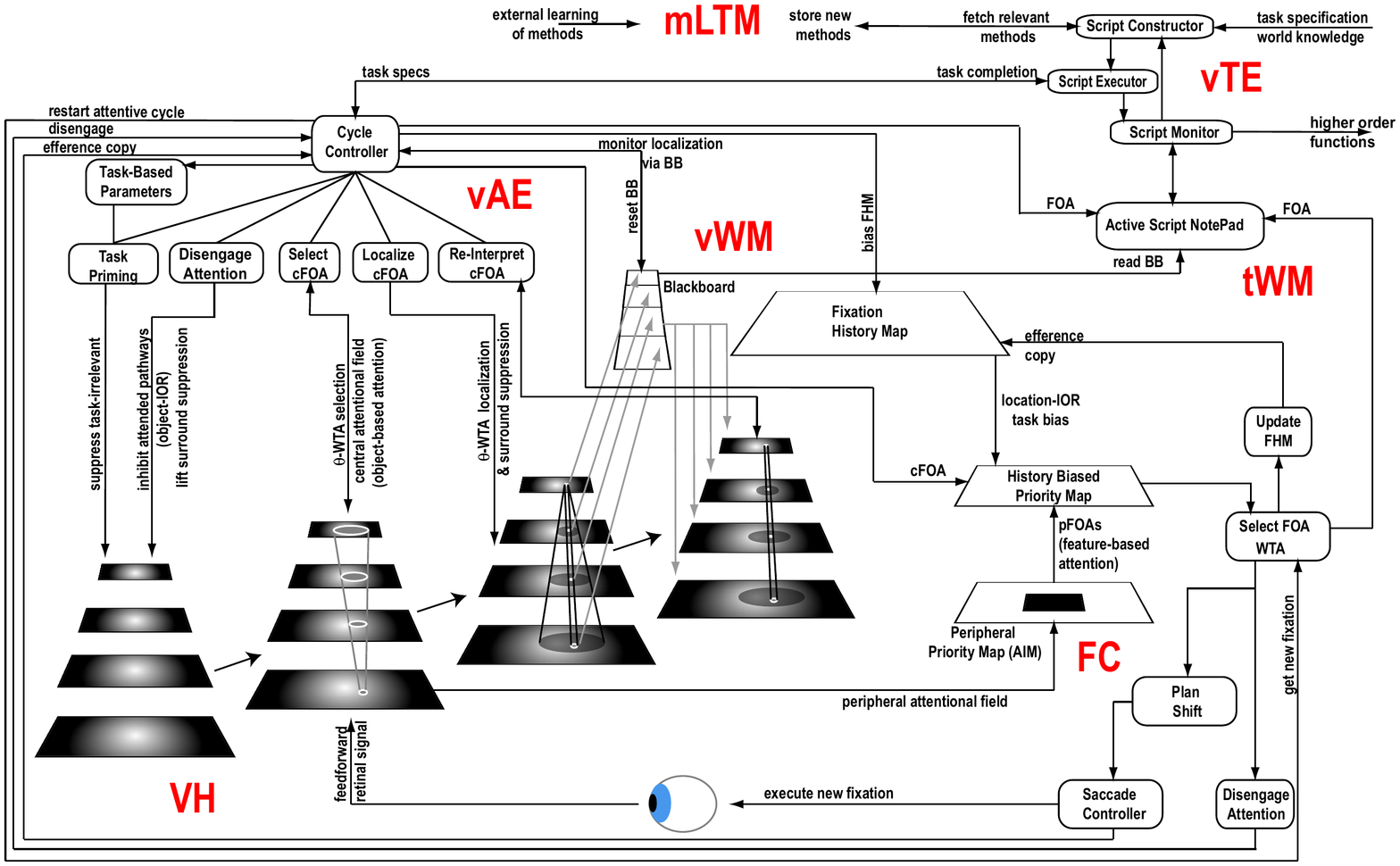}
\par\end{centering}
\caption{\label{fig:Detailed-view-of}Detailed view of the STAR architecture.
. Reproduced from \citep{tsotsos2014cognitive}}
\end{figure*}
\par\end{center}

The Selective Tuning with Cognitive Programs framework allows a very
rich set of visual tasks to be solved, given the correct sequence
of methods is performed. A recent realization of Cognitive Programs
in challenging environments has been presented in \citep{New1} where
an agent is able to successfully play two video games by using a set
of methods to control and tune the Visual Hierarchy and decide on
the next move for the player. 

Nevertheless, some open questions remain, which are (1) how to design
the structure and parameters of the VH so that it can, given the proper
task-biasing / priming, deal with a broad range of visual inputs?
(2) How does one learn the type of tuning that is to applied to VH
for each given task (3) How to create a visual task executive which
is able to appropriately select a set of methods which will accomplish
a visual task. 

It seems that the main questions posed here have to do with control
and with planning solution given some set of tools, as well as fitting
models to a complex data (such as images). It is only natural to proceed
with the recent trends in machine learning which can facilitate the
solution of such problems. The following descriptions are not meant
as very in-depth descriptions of the respective methods, but more
as a high-level overview, elaborating on details as required. Importantly,
we highlight shortcomings that these methods face and suggest solutions
for some. 

\section{Machine Learning}

In this section, we provide an overview of the main methods in machine
learning which are relevant to require some intelligent agent to observe
the world and perform various given complex tasks. This seems like
a very broad subject, certainly one which is yet to be fully solved
(as having this fully solved would mark the start of a general AI).
Yet, notable progress has been made in recent years in machine learning
and pattern recognition. In this short exposition, we mention the
two main methods of interest which we deem relevant to the current
goal of this work.

\subsection{Deep Learning}

Deep learning is certainly not a new field and has its roots set back
in the 1960's. Due to various reasons which are out of the scope of
this work, it has not always been as popular as today and certainly
there are still those that claim that the current hype around it is
exaggerated. A turning point responsible for its current surge in
popularity is the 2012 paper \citep{krizhevsky2012imagenet} which
won the ImageNet \citep{russakovsky2015imagenet} large-scale visual
recognition challenge. This is a massive benchmark for computer-vision
methods where a classifier is required to predict the class of an
object in an image out of a possible 1000 different classes. Significantly
outperforming all other results, the work spurred an avalanche of
follow-ups and modifications, both from an optimization point of view
and of different architectures, as well as theoretical works attempting
to justify the success of such methods over others. To date, it is
rare to see a leading method in computer vision which is not based
on deep learning, be it in the sub-tasks for object recognition, detection
(i.e, localization), segmentation, tracking, 3D-reconstruction, face
recognition, fine-grained categorization and others. Specifically,
deep convolutional neural networks, a certain form of neural nets
which exploits assumptions about the structure of natural images,
is a main class in deep networks. The success of deep learning has
also spread to other media such as audio (e.g, speech recognition),
natural language processing (translation) and other sub-fields involving
pharmaceutical and medical applications, etc. The literature in recent
years on Deep Learning is vast and the reader is encouraged to turn
to it for more in depth information. \citep{schmidhuber2015deep,litjens2017survey}.

The crux of the various deep-learning based methods lie in their need
for massive amounts of supervised data. To obtain good performance,
tens of thousands (sometimes more) of example are required. While
some semi-supervised methods are being suggested, non have approached
the performance of fully supervised ones. This is not to say that
their utility is discarded - on the contrary, we believe that they
will play a major role in the developments of the near future. Additional
issue lies in their current seemingly inherent inability to adjust
to new kinds of data or apply compositions of already learned solutions
to new problems \citep{rosenfeld2018challenging}. Further weakenesses
of deep learning systems are discussed in \citep{shalev2017failures}
and \citep{marcus2018deep}, as well as a discussion about some major
differences between the way humans and machines solve problems \citep{lake2017building}.

\subsubsection{Semi-supervised and Unsupervised Learning}

One variant of machine learning potentially holds some promise to
ameliorate the need for supervision at scale which is required by
methods such as Deep Learning. Such methods attempt to perform learning
by receiving a much smaller amount of supervision. For example, learning
how to distinguish the data into two classes, but doing so by learning
on a datasets where only 10\% is labeled and the rest is not. This
can be done by exploiting observed similarities in the underlying
data and / or assuming some regularities such as smoothness, etc.
An extreme case would be using no labeled data at all, however, as
at some point there will be a task where a system should learn in
a supervised manner, the utility of the unsupervised learning will
be measured by finding how it benefits the supervised learner. Another
form of unsupervised learning is Generative Models, which is able
to produce at test time data points whose properties ideally resemble
those observed at training time, though of course not identical to
them. An example of Semi-supervised Learning is Ladder-Networks \citep{rasmus2015semi},
where an unsupervised loss is added to the network in addition to
the supervised loss. A notable method which has recently gained popularity
are Generative Adversarial Networks (GAN) \citep{goodfellow2014generative},
where two networks constantly compete: the goal of the Generator network
is to generate images which are as realistic as possible in the sense
that they resemble images from the training set and a Discriminator
network whose goal is to tell apart the images from the Generator
and the images from the real dataset. This has quickly evolved to
produce impressive results, a recent one due to \citep{nguyen2016plug},
see Fig. \ref{fig:Output-of-Conditional} for some results. 
\begin{center}
\begin{figure*}
\begin{centering}
\includegraphics[width=0.8\textwidth]{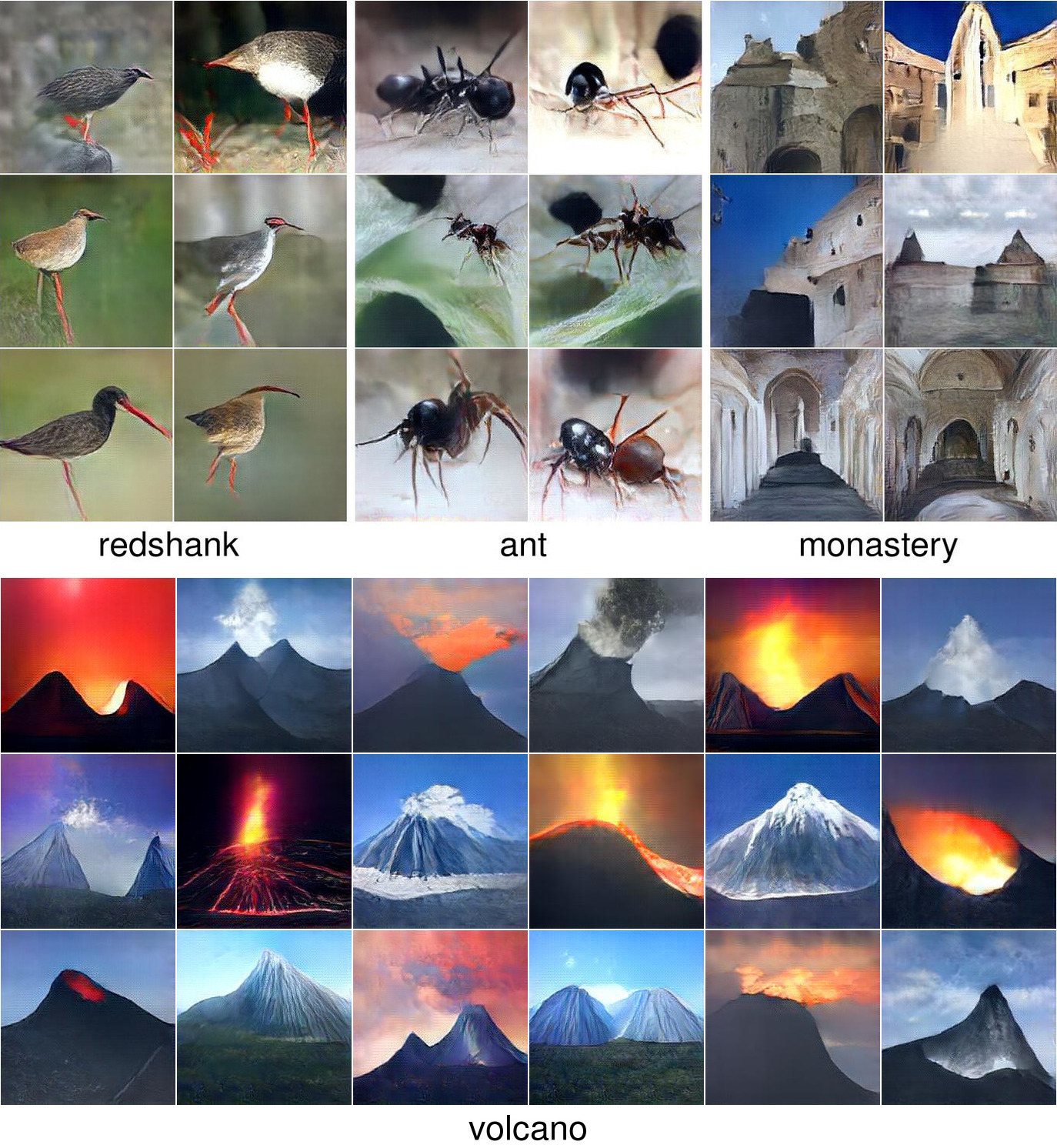}
\par\end{centering}
\caption{\label{fig:Output-of-Conditional}Output of Conditional Generative
Adversarial Network, generating images to resemble certain classes.}

\end{figure*}
\par\end{center}

\subsection{Deep Reinforcement Learning}

Reinforcement Learning (RL) refers to a set of classical and well
studied methods in the field of control systems and artificial intelligence.
The general setting is that of an agent who is supposed to take actions
in some given environment. As a result the agent may encounter new
situations and be given some reward (or penalty). The actions of the
agent may affect the environment. The agent does not necessarily see
the entire environment at all times, rather have access to some input
which is its current observation. Through this loop of act-observe-receive
reward the agent must increase its total future reward. This setting
is very general in the sense that it is limited only by the richness
of the environment and of the agent. For an extreme example, we may
say that the environment is planet Earth and the agent is some human
being or animal. As simulating either of these seems like a virtual
impossibility one can model e.g, robots in closed and well defined
environments, or anything in between. Much research has gone into
making agents which can learn and are able to perform well in various
environments, as well as making robust control systems. RL was also
one of the fields to benefit and regrow in popularity following the
success of deep learning, leading to a new method called Deep Reinforcement
Learning. The first widely known success of this new method has been
published in \citep{mnih2013playing}, where an agent was shown to
be able to learn how to perform well in multiple Atari video games,
outperforming many previous methods. Notably, the system was learned
end-to-end without any input except the raw pixel data and the score
of the game. In some cases, it has even learned to outperform human
players. The reported performance was a result of using a single architecture
(except the number of output variables where games had different number
of possible controls) and set of hyper-parameters. Although there
were many game on which the method performed poorly at the time (and
still does), this was a significant results which lead to others,
such as beating a human expert in the game of GO \citep{silver2016mastering}
which is widely acknowledged as a long standing challenge for the
artificial intelligence community. For a recent overview on this subject,
please refer to \citep{li2017deep}. 

Formally, RL assumes the following setting: an agent may interact
with an environment at each time t by applying an action $a_{t}$,
given an observation $s_{t}$. Note that the entire state of the environment
may not be observed, and in this context the state $s_{t}$ represents
only the observation of the agent - it is all that it can directly
measure. The interaction $a_{t}$ of the agent leads to another state
$s_{t+1}$, where the agent may perform another action $a_{t+1}$,
and so on. A Markov Decision Process (MDP) is defined as an environment
where the probability of the next state is fully determined by the
current state and the action:

\begin{equation}
P_{ss'}^{a}=Pr(s_{t+1}=s'\mid s_{t}=s,a_{t}=a)
\end{equation}

i.e., the probability of state $s'$ following state $s$ after action
$a$. 

Note that this has the Markov property, i.e., that each state is dependent
only on the previous one and not on ones before that. For example,
in a game of Chess, where the entire board is observed as the state,
nothing needs to be known about previous steps of the game to determine
the next move. For each action the agent receives a reward $r_{t}$,
which is a real scalar that can take on any value, be it positive,
negative or zero. Hence, the entire sequence of $n$ actions of an
agent in an environment is 
\begin{equation}
s_{0},a_{0},r_{1},s_{1},a_{1},r_{2},s_{2},a_{2},r_{3},\dots s_{n-1},a_{n-1},r_{n},s_{n}
\end{equation}

Where $s_{n}$ is the terminal reward (win/lose/terminate). The goal
of the agent is to maximize the total future reward: assuming that
the agent performed $n$ steps and at each step received a reward
$r_{t},$the total reward is 
\begin{equation}
R=\sum_{t=1}^{n}r_{t}
\end{equation}

The total \emph{future }reward from time $t$ is 
\begin{equation}
R_{t}=\sum_{i=1}^{n-t}r_{t+i}
\end{equation}

However, as the close future holds less uncertainty, it is common
to consider the \emph{discounted future reward}, that is a reward
which is exponentially decayed over time:

\begin{align}
R_{t} & =r_{t}+\gamma r_{t+1}+\gamma^{2}r_{t+2}+\dots+\gamma^{n-t}r_{n}\\
 & =r_{t}+\gamma(r_{t+1}+\gamma(r_{t+2}+\dots))\\
 & =r_{t}+\gamma R_{t+1}\label{eq:discounted}
\end{align}

The strategy that the agent uses to determine the next action is called
a \emph{policy}, and is usually denoted by $\pi$. A good policy would
maximize the discounted future reward. A value-action function $Q$
is defined as a function which assigns the maximum discounted future
reward for an action $a_{t}$ performed at a state $s_{t}$:

\begin{equation}
Q(s_{t},a_{t})=\max R_{t+1}
\end{equation}

Given this function, the optimal policy can simply choose for each
state $s$ the action $a$ which maximizes $Q$:

\begin{equation}
\pi(s)=\argmax_{a}Q(s,a)
\end{equation}

From Eq. \ref{eq:discounted}the following relation holds: 

\begin{equation}
Q(s,a)=r+\gamma\max_{a'}Q(s',a')
\end{equation}

Meaning that if we find a function $Q$ for which the above holds,
we can use it to generate an optimal policy. For a discrete number
of states and actions, a simple method known as \emph{value iteration}
is know converge to the optimal policy \citep{sutton1998reinforcement},
given that each state/action pair is visited an infinite number of
times. This is simply implemented as continuously updating $Q$, until
some stopping criteria is met:

\begin{equation}
Q(s_{t},a_{t})\leftarrow Q(s_{t},a_{t})+\alpha[r+\gamma\max_{a_{t+1}}Q(s_{t+1},a_{t+1})-Q(s_{t},a_{t})].
\end{equation}

Such methods can work well for a finite number of states and actions.
However, for many interesting environments it is challenging to define
the states as a discrete set, and doing so naively would result in
an exponential number. For examples, if the task is to play a video
game, while the available actions form a small set, enumerating the
number of possible stimuli would easily lead to intractable numbers,
that is, all possible combinations of pixel values on the screen. 

With this in mind, we turn to Deep Reinforcement Learning. Here, instead
of representing each state explicitly as some symbol in a large set,
a neural network is learned to predict the $Q$-value from the state,
by being applied directly to each input frame (or set of a few consecutive
ones to capture motion). Hence the state is represented implicitly
by the network's weights and structure. This allows the agent to learn
how to act in the environment without having privileged knowledge
about its specific inner workings. There are many variants and improvements
on this idea, though the basic setting remains the same. In what follows,
we highlight some of the challenges and shortcomings of the current
approaches of Deep RL.

\subsubsection{Efficient Exploration}

A very big challenge currently holding back RL methods is the huge
exploration space that should potentially be sought in order to produce
a good policy. This is a chicken and egg problem of sorts: exploration
is needed to find out a good policy and a good policy is required
to be able to do sufficient exploration; consider even a simple game
such as Atari Breakout, where the player is able to move a paddle
left or right and hit the ball so it doesn't fall off the bottom of
the screen. If nothing else is known, it would take some amount of
exploration to find out the paddle should bounce the ball to avoid
losing. Before that, the agent will probably start off just moving
randomly to the left and right. As the rewards of the game are sparse,
it will not be until the agent encounters its first reward that it
will be able to update its policy. For this reason, it will take many
iterations until it can start learning how to act to avoid losing
quickly. Only then can it continue to explore further states of the
game, which could not even be reached if it had not passed the very
first steps of hitting the ball. The $\epsilon$greedy strategy somewhat
improves on this problem by choosing a random move with a probability
of $\epsilon$, a hyper-parameter which is usually decayed over time
as the system learns. This helps getting out of local minima in the
exploration space, though the general problem described here is certainly
not solved. 

\subsubsection{Exploration vs Representation}

The problem of exploration is exacerbated for the case of Deep RL.
In a discrete search space, each state is well recognized once encountered.
When the state space is represented implicitly by a deep network,
the evolution of the Q function is tied with the representation of
the environment by the Deep network. This means that updating the
Q function can lead to unstable results. One strategy to address this
is by reducing the frequency in which the network which chooses the
next action is updated . We suggest here a couple of additional strategies:
\begin{itemize}
\item A strong visual representation: the visual system of human beings
is a strong one and is able to represent stimuli very robustly, owing
both to evolution and learning from prior experience. An agent usually
learns the visual representation of the environment from scratch.
Certainly, being able to robustly represent the observations right
from the start would allow the agent to focus more on planning and
less on learning the representation. Nevertheless, the representation
may continue evolving as the agent encounters new situations. One
way to allow this is to use as a starting point a pre-trained visual
representation, be it in a supervised or unsupervised manner, and
adapt it as needed for the task. 
\item Symbolic representation: allowing the agent to group observations
into equivalence classes by assigning symbols or compact representations
to them would allow policies to be learned more efficiently and probably
converge to a higher level of performance. This can be done in an
implicit manner by attempting to cluster the representation of the
environment into few informative clusters which carry the maximal
information with respect to the task. More explicitly, the observation
can be somehow parsed into objects, background, possibly other agents,
etc. The representation of the scene will then be made up of the properties
(speed, location, state) of the constituents of the scene. While the
latter would probably carry more meaning (and presumably lead to higher
performance), it seems hard to do so in a purely data-driven approach
without external knowledge about the world. 
\end{itemize}

\subsubsection{Prior and External Knowledge}

A child is able to learn how to play a game reasonably well within
a few minutes (a few tens of thousands of frames). Current methods
require many millions of frames to do so, if they succeed at all.
Why is this so? Besides the reasons stated above, we claim that additional
forms of prior experience are useful. 

One form of experience is having solved tasks in the past which may
be related to the current task. Indeed, this has been recently shown
to be effective in \citep{parisotto2015actor} where a single network
learns to mimic the behaviour of multiple expert networks, each of
which was pre-trained on a single tasks. Thus the new network represents
simultaneously the knowledge to solve all of the learned tasks in
a relatively compact manner. In most cases, such a network was shown
to learn new tasks much faster than a randomly initialzied version
as well as converge in a more stable manner.

World knowledge also plays a major role in understanding a new situation.
The factual knowledge we gain from experience, if written as a list
of many different facts and rules, would probably make a very long
one. Here are a few examples:
\begin{itemize}
\item An intuitive understanding of Newtonian Physics - even children understand
that object tend to continue in their general direction, tend to fall
down after going up, may move if pushed by some external force, etc. 
\item Relations and interactions between objects: doors may require keys
to be opened
\item Survival: falling off a cliff is usually a bad idea; if an opponent
comes your way, you'd better avoid it or terminate it
\item General facts: roses are red. Violets are blue. Gold gives you points. 
\end{itemize}
It is difficult to imagine how all of this is learned and stored in
our brains and how the relevant facts come into play in the abundance
of different situations that we encounter. Being able to effectively
utilize such a vast knowledge-base about the behavior of the world
would no doubt aid intelligent agents in many environments. Attempts
at using external knowledge to aid tasks have already been made in
Computer Vision for image captioning and Visual Question Answering
(VQA) \citep{wu2016image}, Zero-Shot Learning (ZSL) \citep{frome2013devise}
and in general to gain knowledge about unseen objects or categories
by comparing their detected attributes to those of known ones \citep{farhadi2009describing}.
Such world-knowledge is collected either by data-driven approches
such as word2vec (a learned vector space representation of words)
\citep{mikolov2013distributed} or word relation graphs (WordNet \citep{miller1995wordnet}),
datasets collected manually or by scanning online knowledge collections
such as Wikipedia, such as ConceptNet \citep{speer2012representing}. 

Such collections of linguistic and factual knowledge can certainly
help an agent quickly reason about its surrounding environment - \emph{only
if it is able to link its observations to items in the knowledge base}.
It is interesting to ask how a person acquires such knowledge in the
first years of his/her lifetime, through an experience which is quite
different that simply being explosed to millions of online articles.
Somehow a collection of useful facts and rules is picked up from experience
despite being drowned in a pool of distracting and noisy signals. 

Some recent work by \citep{dubey2018investigating} has demonstrated
that prior knowledge is quite critical to the success of humans in
simple games. The work devises a few ways to remove the semantics
from gameplay by replacing graphical elements in the game by semantically
meaningless ones. For example, switch each piece of texture in the
game to a random one (but do so consistently). This makes the game
screen appear meaningless to the human observer. The performance of
humans in such modified games dropped significanlty while that of
the tested machine-learning based method remained the same. Another
type of modification was switching elements with elements of different
meaning. An example is replacing the appearance of a ladder to be
climbed to a column of flames, or transposing the screen so gravity
appears to work sideways. Though there is a one-to-one translation
between the original and modified version of the game, human players
did much worse on these semantically modified examples, and on others.
This demonstrates the heavy reliance humans have on prior knowledge.
In this context, learning a game from scratch without prior knowledge
is ``unfair'' for machine-learning methods. 

Nevertheless, such knowledge bases still do not account for an intuitive
physical understanding, which seems to require some other type of
experience. Such knowledge can either be pre-injected into the agent
but, as children do not come equipped with such knowledge, we believe
that the agent should learn the rules of physical interactions from
its own experience or observations. An interesting attempt at this
direction can be seen in \citep{agrawal2016learning} where robots
gain a reportedly ``intuitive'' understanding of physical interactions
by attempting to perform simple tasks on objects such as moving them
around.

\subsubsection{High Level Reasoning and Control}

Planning can be performed at several levels of granularity. Certainly,
a human being or animal does not think in terms of the force that
needs to be applied by each of the muscels in order to pick up some
object. It rather seems that plans are made at a higher level of abstraction
and some process then breaks them down to motor commands and everything
that is required for them to be carried out. The motor commands can
also be grouped into logical units above the most basic ones, such
as ``fully stretch out left arm'', which is only then translated
to low level commands. Newborns are not able to control their limbs
and fingers immediately, but over time they acquire this ability and
perform tasks with seamless movements, usually dedicating little or
no concious thought to the movement of muscles. Similarly, exploration
which goes on early on in the ``life'' of an agent should allow
the agent to learn how to perform simple and common actions and store
these as routines to be later used in more elaborate plans. End-to-end
learning of motor policies from raw pixel data is attempted in \citep{levine2016end}
.

The above was only the simplest level of high-level control. Further
advances would requires strategic thinking in terms of long-range
goals and actions. We claim that this cannot be done effectively without
first obtaining a hierarchy of basic control over the agents actions
and being able to predict quite reliably their immediate future effect. 

\bibliographystyle{ieeetr}
\bibliography{cognitivePrograms1}

\end{document}